\documentclass{article}

\PassOptionsToPackage{numbers, compress}{natbib}

\usepackage[final]{neurips_2019}

\usepackage[utf8]{inputenc} 
\usepackage[T1]{fontenc}    
\usepackage{hyperref}       
\usepackage{url}            
\usepackage{booktabs}       
\usepackage{amsfonts}       
\usepackage{nicefrac}       
\usepackage{microtype}      

\usepackage{float}

\usepackage{amsfonts}
\usepackage{amsmath}
\usepackage{amsthm}
\usepackage{amssymb}

\theoremstyle{plain}
\newtheorem{theorem}{Theorem}[section]
\newtheorem{lemma}{Lemma}

\theoremstyle{definition}
\newtheorem{definition}{Definition}
\newtheorem{proposition}{Proposition}

\usepackage{graphicx}
\usepackage{subcaption}
\graphicspath{{figures/}{../figures/}}

\usepackage{subfiles}
\def\biblio{\bibliographystyle{plainnat}\bibliography{references}}  

\usepackage{xcolor}

\usepackage{tikz}

\usepackage[ruled]{algorithm2e}

\usepackage{mathtools}

\DeclareMathOperator*{\argmax}{arg\,max}

\usepackage{wrapfig}

\title{Categorized Bandits}

\author{%
  Matthieu Jedor \\
  CMLA, ENS Paris-Saclay \& Cdiscount\\
  \texttt{jedor@cmla.ens-cachan.fr} \\
   \And
   Jonathan Lou\"{e}dec \\
   Cdiscount \\
   \texttt{jonathan.louedec@cdiscount.com} \\
   \And
   Vianney Perchet \\
   CMLA, ENS Paris-Saclay \& Criteo AI Lab \\
   \texttt{perchet@cmla.ens-cachan.fr} \\
}

\begin{document}

\def\biblio{}

\maketitle

\begin{abstract}
We introduce a new stochastic multi-armed bandit setting where arms are grouped inside ``ordered'' categories. The motivating example comes from e-commerce, where a customer typically has a greater appetence for items of a specific well-identified but unknown category than any other one. We introduce three concepts of ordering between categories, inspired by stochastic dominance between random variables, which are gradually weaker so that more and more bandit scenarios satisfy at least one of them. We first prove instance-dependent lower bounds on the cumulative regret for each of these models, indicating how the complexity of the bandit problems increases with the generality of the ordering concept  considered. We also provide algorithms  that fully leverage the structure of the model with their associated theoretical guarantees. Finally, we have conducted an  analysis on real data to highlight that those ordered categories actually exist in practice.
\end{abstract}

\section{Introduction}

In the multi-armed bandit problem, an agent has several possible decisions, usually referred to as ``arms'', and chooses or ``pulls'' sequentially one of them at each time step. This generates a sequence of rewards and the objective is to maximize their cumulative sum. The performance of a learning algorithm is then evaluated through the ``regret'', which is the difference between the cumulative reward  of an oracle (that knows the best arm in expectation) and the cumulative reward of the algorithm. There is a clear trade-off  arising between gathering information on uncertain arms (by pulling them more often) and using this information (by choosing greedily the best decision so far). This tradeoff is usually called ``exploration vs exploitation''. Although originally introduced for adaptive clinical trials~\cite{thompson1933likelihood}, multi-armed bandits now play an important role in recommender systems~\cite{li2010contextual}. However, the traditional bandit model (see \citet{bubeck2012regret} for more details and variants) must be adapted to specific applications to unleash its full power.
 
Consider for instance e-commerce. One of the core optimization problem is to decide which products to recommend, or display, to a user landing on a website, in the objective of maximizing the click-through-rate or the conversion rate. Arms of recommender systems are the different products that can be displayed. The number of products, even if  finite, is prohibitively huge as the regret, i.e.\ the learning cost, typically scale linearly with the number of arms. So agnostic bandit algorithms  take too much time to complete their learning phase. Thankfully, there is an inherent structure behind a typical catalogue: products are gathered into well defined categories. As customers are generally interested in only one or a few of them, it seems possible and profitable to gather information across products to speed up the learning phase and, ultimately, to make more refined recommendations. 
 
\paragraph{Our results} We introduce and study the idea of \emph{categorized bandits}. In this framework, arms are grouped inside known categories and we assume the existence of a partial yet unknown order between categories. We aim at leveraging this additional assumption to reduce the linear dependency in the total number of arms. 
We present three different partial orders over categories inspired by different notions of stochastic dominance between random variables. We considered gradually weaker notions of ordering in order to cover more and more bandit scenarios. On the other hand, the stronger the assumption, the more ``powerful'' the algorithms are, i.e.\ their regret is smaller. Those assumptions are motivated and justified by real data gathered on the e-commerce website Cdiscount. We first prove asymptotic instance-dependent lower bounds on the cumulative regret for each of these models, with a special emphasis on how the complexity of the bandit problems increases with the generality of the ordering concept considered. We then proceed to develop two generic algorithms for the categorized bandit problem that fully leverage the structure of the model; the first one is devised from the principle of optimism in the face of uncertainty~\cite{auer2002finite} when the second one is from the Bayesian principle~\cite{thompson1933likelihood}. Finite-time instance-dependent upper bounds on the cumulative regret are provided for the former algorithm. Finally, we conduct numerical experiments on different scenarios to illustrate both finite-time and asymptotic performances of our algorithms compared to algorithms either agnostic to the structure or only taking it partly into account.

\paragraph{Related works} 
The idea of clustering is not novel in the bandit literature~\cite{nguyen2014dynamic, bresler2014latent, gentile2014online, korda2016distributed, li2016graph} yet they mainly focus on clustering users based on their preferences. \citet{li2016collaborative} extended these work to the clustering of items as well. \citet{katariya2018adaptive} considered a problem where the goal is to sort items according to their means into clusters. Similar in spirit are bandit algorithms for low-rank matrix completion~\cite{zhao2013interactive, kawale2015efficient, katariya2016stochastic}. 
\citet{maillard2014latent} studied a multi-armed bandit problem where arms are partitioned into latent groups. \citet{valko2014spectral} and \citet{kocak2014spectral} proposed algorithms where the features of items are derived from a known similarity graph over the items. However, none of these works consider the known structure of categories in which the items are gathered.
The model fits in the more general structured stochastic bandit framework i.e.\ where expected reward of arms can be dependent, see e.g., \cite{lattimore2014bounded, degenne2016combinatorial, abe1999associative,kwon2016gains,perrault2019finding}. More recently, \citet{combes2017minimal} proposed an asymptotically optimal algorithm for structured bandits relying on forced exploration (similarly to \cite{Lattimore17}) and a tracking mechanism on the number of draws of sub-optimal arms. However, these approaches forcing exploration are too conservative as the linear dependency only disappears asymptotically.
There exist two other ways to tackle the bandit problem with arms grouped inside categories. The first one could rely on tree search methods, popularized by the celebrated \textsc{uct} algorithm~\cite{kocsis2006bandit}. 
Alternative hierarchical algorithms~\cite{coquelin2007bandit} could also be used. The second one could be linear bandits~\cite{dani2008stochastic, rusmevichientong2010linearly, abbasi2011improved} where we introduce a ``categorical'' feature that indicates in which category the arm belongs. However, these approaches are also not satisfactory as they do not leverage the full structure of the problem.

\section{Model}

We now present the variant of the multi-armed bandit model we consider.  As usual, a decision maker sequentially selects (or pulls) an arm at each time step $t \in \{1,\ldots,T\}=:[T]$ . As motivated in the introduction, the total number of possible arms can be prohibitively large, but we assume that this large number of arms are grouped in a small number $M$ of categories. For the sake of presentation, we are going to assume that each category has the same number of arms $K$, yet all of our assumptions and results immediately generalize to different number of arms. We emphasize again that the $M$ categories of $K$ arms each form a known partition of the set of arms (of cardinality $MK$).
At time step $t \in [T]$, the agent selects a category $C_t$ and an arm $A_t \in C_t$ in this category. This generates a reward $X^{C_t}_{A_t} =\mu^{C_t}_{A_t} + \eta_t$ where $\eta_t$ is some independent 1 sub-Gaussian white noise and  $\mu^m_k$ is the unknown expected reward of the arm  $k$ of category $m$. For notational convenience, we will assume that arms are ordered inside each category, i.e.\ $\mu^m_1 > \mu^m_2 \geq \dots \geq \mu^m_{K-1} > \mu^m_K$ for all category $m$ and that category 1 is the best category, with respect to a partial order defined below.\footnote{To be precise, since the order is only partial, some categories  might not be pairwise comparable, but we assume that the optimal category is comparable to, and dominates, all the others.} We stress out  that, in the partial orders we consider,  the maximum of $\mu^m_k$ over $m$ and $k$ is necessarily  $\mu_1^1$.
As in any multi-armed bandit problem, the overall objective of an agent  is to maximize her expected cumulative reward  until time horizon $T$ or identically, to minimize her expected cumulative regret $\mathbb{E}[R_T] = T \mu_1^1 - \mathbb{E}[ \sum_{t=1}^T \mu_{A_t}^{C_t}]$, or equivalently,  
$\mathbb{E}[R_T] = \sum_{m, k} \Delta_{m,k} \, \mathbb{E}[N_k^m(T)]$, 
where $\Delta_{m, k} \coloneqq \mu_1^1 - \mu_k^m$ is the difference, usually called ``gap'', between the expected rewards of the best arm and the $k$\textsuperscript{th} arm of category $m$ and $N_k^m(t) \coloneqq \sum_{s < t} \mathbf{1}\{ C_s = m, A_s = k \}$ denotes the number of times this arm  has been pulled up to (not including) time step $t$. 

\paragraph{Relations of dominance} The main assumption to leverage is that the set of categories is partially ordered with a unique maximal element. Those partial orders are quite similar to the standard ones induced by stochastic dominance \cite{hadar1969rules, bawa1975optimal} over random variables. 
We  are going to consider three notions of dominance (inducing three different partial orders) that are gradually weaker so that the bandit setting is more and more general. Consequently, the regret should be higher and higher. 

\begin{definition} Let $\mathcal{A} = \{\mu_1^{\mathcal{A}}, \ldots, \mu_K^{\mathcal{A}}\} \subset \mathbb{R}$ and $\mathcal{B} = \{\mu_1^{\mathcal{B}}, \ldots, \mu_K^{\mathcal{B}}\} \subset \mathbb{R}$ be a pair of categories, 
\vspace{-8pt}
\begin{description}
\item[Group-sparse dominance]  $\mathcal{A}$ group-sparsely dominates $\mathcal{B}$, denoted by $\mathcal{A} \succeq_s \mathcal{B}$, if each element of $\mathcal{A}$ are non-negative and at least one is positive, and each element of $\mathcal{B}$ are non-positive, i.e.,
$\displaystyle \max_{k \in [K]} \mu_k^{\mathcal{A}}  > \min_{k \in [K]} \mu_k^{\mathcal{A}} \geq 0  \geq \max_{k \in [K]} \mu_k^{\mathcal{B}}\,.$
\item[Strong dominance] $\mathcal{A}$ strongly dominates  $\mathcal{B}$, denoted by $\mathcal{A} \succeq_0 \mathcal{B}$, if each element of $\mathcal{A}$ is bigger than any element of $\mathcal{B}$, i.e.,
$\displaystyle \min_{k \in [K]} \mu_k^{\mathcal{A}} \geq \max_{k \in [K]} \mu_k^{\mathcal{B}}\,.$
\item[First-order dominance]  $\mathcal{A}$ first-order dominates  $\mathcal{B}$, denoted by $\mathcal{A} \succeq_1 \mathcal{B}$, if 
$\displaystyle \sup_{x \in \mathbb{R}} F_{\mathcal{A}}(x) -  F_{\mathcal{B}}(x) \leq 0 \,,$
where $F_\mathcal{A}(x) = \frac{1}{K} \sum_{k=1}^{K} \mathbf{1}\{ \mu_k^{\mathcal{A}} \leq x \}$ is the cumulative distribution function of a uniform random variable over $\mathcal{A}$ (and similarly for $\mathcal{B}$).
\end{description}
\end{definition}

The first notion of dominance is inspired by the classical (group-)sparsity concept in machine learning, that already emerged in variants of multi-armed bandits~\cite{kwon2017sparse, bubeck2013bounded}. It is quite a strong assumption as it implies the knowledge of a threshold\footnote{This threshold is fixed at 0 for convenience, but it could have any value.} between two categories. The second notion weakens this assumption as the threshold  is unknown. The third notion is even weaker.
The second and third notions of dominance are similar to the zeroth (also called strong) and first-order of stochastic dominances between two random variables respectively uniform over $\mathcal{A}$ and $\mathcal{B}$. Hence, the three concepts of dominance immediately generalize to categories with different number of elements, with the very same definitions. Furthermore, one can weaken even more the dominance, e.g. introducing a second-order variant, but we will not consider it in this paper.

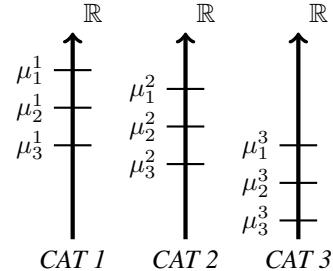
\begin{wrapfigure}[12]{r}{0.36\linewidth}
\centering
\vspace{-11pt}
\begin{tikzpicture}
\draw[black, ultra thick, ->] (0,-0.25) node[anchor=north]{\emph{CAT 1}} -- (0,2.5) node[anchor=south west]{$\mathbb{R}$};
\draw[black, thick] (-0.25,2) node[anchor=east]{$\mu^1_1$} -- (0.25,2);
\draw[black, thick] (-0.25,1.5) node[anchor=east]{$\mu^1_2$} -- (0.25,1.5);
\draw[black, thick] (-0.25,1) node[anchor=east]{$\mu^1_3$} -- (0.25,1);

\draw[black, ultra thick, ->] (1.5,-0.25) node[anchor=north]{\emph{CAT 2}} -- (1.5,2.5) node[anchor=south west]{$\mathbb{R}$};
\draw[black, thick] (1.25, 1.75) node[anchor=east]{$\mu^2_1$} -- (1.75, 1.75);
\draw[black, thick] (1.25,1.25) node[anchor=east]{$\mu^2_2$} -- (1.75,1.25);
\draw[black, thick] (1.25,0.75) node[anchor=east]{$\mu^2_3$} -- (1.75,0.75);

\draw[black, ultra thick, ->] (3,-0.25) node[anchor=north]{\emph{CAT 3}} -- (3,2.5) node[anchor=south west]{$\mathbb{R}$};
\draw[black, thick] (2.75,1) node[anchor=east]{$\mu^3_1$} -- (3.25,1);
\draw[black, thick] (2.75,0.5) node[anchor=east]{$\mu^3_2$} -- (3.25,0.5);
\draw[black, thick] (2.75,0) node[anchor=east]{$\mu^3_3$} -- (3.25,0);
\end{tikzpicture}

\caption{Illustration of dominances} 
\label{fig:dom}
\end{wrapfigure}

\paragraph{Example} To illustrate the concepts of dominance, we have represented, in Figure~\ref{fig:dom},   3 categories of 3 arms each.
It can be easily checked that, for the first-order dominance, 
$\text{\emph{CAT 1}} \succeq_1 \text{\emph{CAT 2}} \succeq_1 \text{\emph{CAT 3}}$
as, if they have the same number of elements, $\mathcal{A}$ first-order dominates $\mathcal{B}$ if the $k$\textsuperscript{th} largest elements of $\mathcal{A}$ is greater than the $k$\textsuperscript{th} largest element of $\mathcal{B}$, for any $k$.
Moreover, for the strong dominance, $\text{\emph{CAT 1}} \succeq_0 \text{\emph{CAT 3}}$
since the worst mean of \emph{CAT 1} is higher than the best mean of \emph{CAT 3}. Moreover, if this common value was known, then the dominance would even be group-sparse.
 

\begin{lemma}
Let $\mathcal{A}_1, \ldots, \mathcal{A}_M$ be finite categories. If there is a category $\mathcal{A}^*$ that dominates all  the other ones for any of the partial orders defined above, then $\mathcal{A}^*$ contains the maximal element of the union $\mathcal{A}_1 \cup \mathcal{A}_2 \cup \ldots \cup \mathcal{A}_M$.
Moreover, if $\mathcal{A}$ group-sparsely dominates $\mathcal{B}$, then the dominance  also holds in the strong sense. Similarly, if $\mathcal{A}$ strongly dominates $\mathcal{B}$, then the dominance also holds in the first-order sense. 
\end{lemma}


\subsection{Empirical evidence of dominance}

\begin{figure}[t]
\centering
\begin{subfigure}{0.435\textwidth}
\centering
\vspace{3pt}
\resizebox{\columnwidth}{!}{%
\begin{tabular}{llll}
\multicolumn{1}{c}{\bf CAT 1}  &\multicolumn{1}{c}{\bf CAT 2}
&\multicolumn{1}{c}{\bf CAT 3}
&\multicolumn{1}{c}{\bf CAT 4} \\
\hline \\
0.0133 &0.0140 &0.0089 &0.0069 \\
0.0114 &0.0088 &0.0086 &0.0063 \\
0.0108 &0.0083 &0.0078 &0.0053 \\
0.0107 &0.0082 &0.0056 &0.0051 \\
0.0096 &0.0078 &0.0052 &0.0051 \\
0.0095 &0.0078 &0.0050 &0.0044 \\
0.0088 &0.0078 &0.0049 &0.0042 \\
0.0086 &0.0077 &0.0047 &0.0041 \\
0.0084 &0.0076 &0.0042 &0.0040 \\
0.0080 &0.0074 &0.0041 &0.0038 \\
\end{tabular}}
\caption{Click-through rates}
\label{table:dom}
\end{subfigure}
\begin{subfigure}{0.555\textwidth}
\centering
\vspace{-3pt}
\includegraphics[width=1\linewidth]{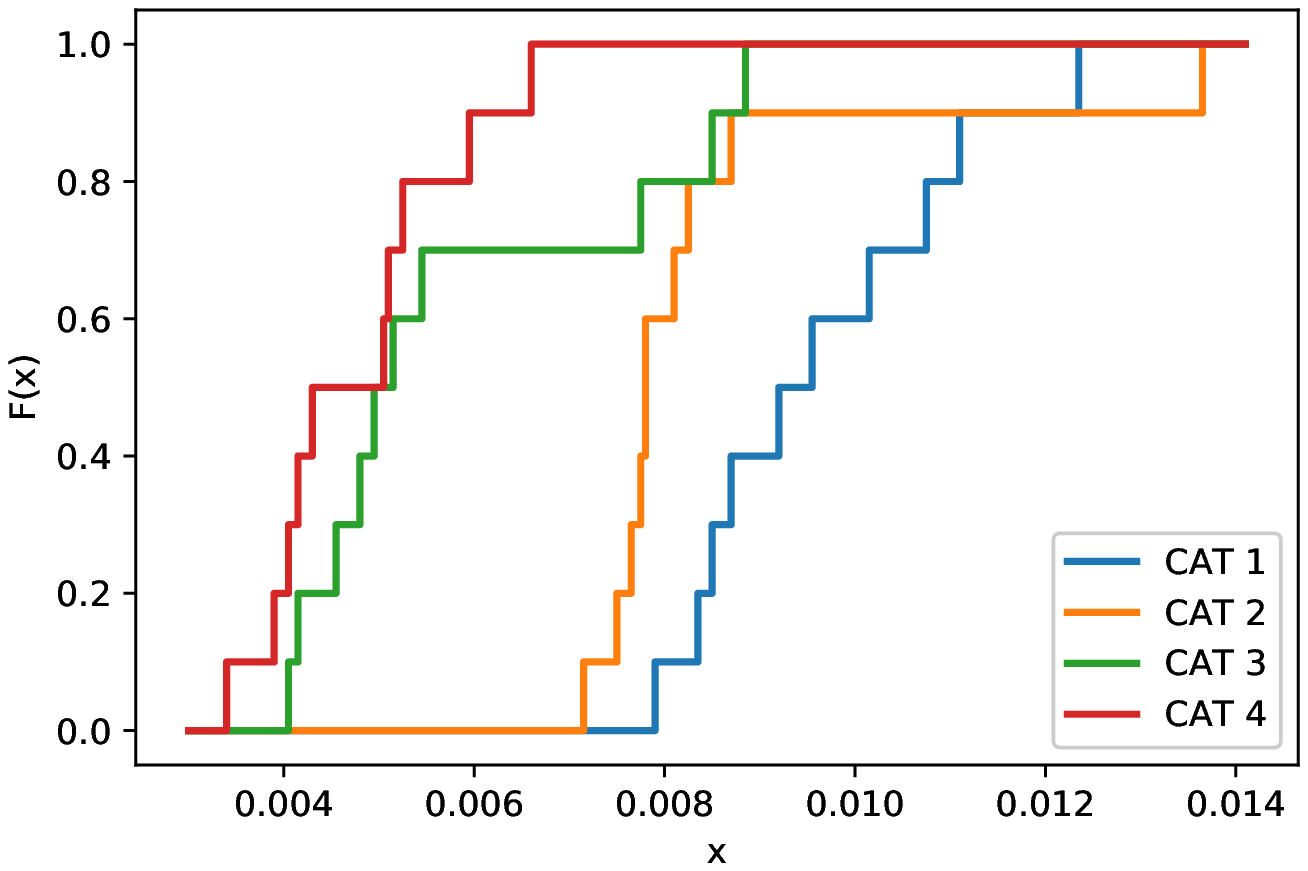}
\caption{Cumulative distribution functions}
\label{fig:DomREal}
\end{subfigure}

\caption{Illustrations of dominance on a real dataset}
\label{fig:empirical evidence}
\end{figure}

We illustrate these assumptions on a real dataset. We have collected the CTR of products in four different categories over  one month on the e-commerce website Cdiscount, one of the leading e-commerce companies in France, gathered in Table~\ref{table:dom}. \emph{CAT} 1 to 3 are three of the largest categories\footnote{For privacy reason, the exact content of the different categories cannot be revealed.} in terms of revenue while \emph{CAT 4} is a smaller category.
The following dominances can be highlighted.
\begin{description}
\item[Strong dominance] \emph{CAT 1} strongly dominates \emph{CAT 4} as its minimum CTR is 0.008 compared to  the maximum CTR of 0.0069 of the other. 
Similarly, \emph{CAT 2} strongly dominates \emph{CAT 4}.
\item[First-order dominance] \emph{CAT 2}  first-order dominates  \emph{CAT 3} as the CTR of each line of the second column are bigger than those of the third column. This dominance is not strong as 0.0074 is smaller than 0.0089. \emph{CAT 3} first-order but not strongly dominates \emph{CAT 4}. 
\item[Uncomparable categories] \emph{CAT 1} and \emph{CAT 2} are not comparable with respect to any  partial order.
\end{description}
Notice that, had the first item of \emph{CAT 2} performed only 5\% worse than observed,\footnote{The CTR of the best item of \emph{CAT 2} is so higher than the second one, we could expect it is actually an outlier, i.e., an artefact of the choice of that specific month and category.} then \emph{CAT 1} would have been optimal with respect to the first-order dominance. So even if the dominance assumption is not satisfied during that specific month,  assuming it would still give good empirical results. 
The relations of dominance can be easier to determine based on the representation of the associated cdf of Figure~\ref{fig:DomREal}. As the cdf of the random variable uniform on \emph{CAT 4} is, pointwise, the biggest one, this means that this category is first-order dominated by all the other ones. Moreover, it reaches 1 while the cdf of \emph{CAT 1} and \emph{CAT 2} are still at 0. This implies that the dominance of these two categories is even strong. This analysis motivates and validates our assumption.

\section{Lower bounds}

In this section, we provide lower bounds on the regret that any ``reasonable'' algorithm (the precise definition is given below) must incur in a  multi-armed bandit problem, where arms are grouped into partially ordered categories (with a dominating one). To simplify the exposition, we assume here that noises are drawn from Gaussian distribution with unit variance. 
The class of  algorithms we consider are consistent~\cite{lai1985asymptotically} with respect to a given a class of possible bandit problems $\mathcal{M}=\big\{ \mu = (\mu_1,\ldots,\mu_{MK}) \in \mathbb{R}^{MK}\big\}$. We recall that an algorithm is  consistent with $\mathcal{M}$ if, for any admissible  reward vector $\mu \in \mathcal{M}$ and any parameter $\alpha \in (0, 1]$, the regret of that algorithm is asymptotically negligible compared to $T^\alpha$, i.e.,
$\displaystyle \sup_{\alpha \in (0,1)} \limsup_{T \to \infty} \frac{\mathbb{E}_{\mu}\left[R_T\right] }{T^\alpha} = 0 \,.$
\citet{graves1997asymptotically}  proved that any algorithm consistent with $\mathcal{M}$ has a regret scaling at least logarithmically in $T$, with a leading constant $c_\mu$ depending on $\mu$ (and $\mathcal{M}$) i.e.,
$\displaystyle \liminf_{T \to \infty} \frac{\mathbb{E}_{\mu}\left[R_T\right]}{\log(T)} \geq c_\mu \,;$
moreover, $c_\mu$ is the solution of some auxiliary optimization problem. In our setting, it rewrites  as
\begin{equation*}
   \textstyle c_\mu = \min_{N \geq 0} \sum_{m,k} N_k^m \Delta_{m,k} \quad \textrm{subject to} \quad  \sum_{m,k} N_k^m \left( \mu_k^m - \lambda_k^m \right)^2 \geq 2, \forall \, \lambda \in \Lambda(\mu) \,,
   \label{eq:lower_cond}
\end{equation*}
where $\Lambda(\mu) = \left\{ \lambda \in \mathcal{M} ; \mu_1^1 =  \lambda_1^1, \lambda_1^1 < \max_{m,k} \lambda_k^m \right\}$. We point out that the assumption of dominance is hidden in the class of bandit problem $\mathcal{M}$. 
In the remaining and with a slight abuse of notation, we are going to call an algorithm consistent with a dominance assumption if it is consistent with the set of all possible vectors of means satisfying this dominance assumption.

\paragraph{Group-sparse dominance} In this case, the above optimization problem has a closed-form solution.
\begin{theorem}\label{TH:group}
An algorithm  consistent with the group-sparse dominance satisfies
$\displaystyle c_\mu = \sum_{k = 2}^K \frac{2}{\Delta_{1, k}} \,.$
\label{th:lower_sparse}
\end{theorem}

The proof of this result (and the subsequent ones) is postponed to the Appendix. This lower bound indicates that all arms in the optimal category (and only those) should be pulled a logarithmic number of times, hence the regret should only scale asymptotically linearly in the number of arms in the optimal category instead of linearly with the total number of arms.
We want to stress out here that Theorem \ref{TH:group} might have a misleading interpretation. Although the asymptotic regret scales with $K$ and independently of $M$, the finite-stage minimax regret is still of the order of $\sqrt{MKT}$, as with usual bandits. This is simply because the lower-bound proof \cite{bubeck2012regret} of the standard multi-armed bandit case uses set of parameters of the form $(0,\ldots, 0, \varepsilon, 0, \ldots, 0)$ which respect the group-sparse assumption. As a result, the asymptotic lower bound of Theorem \ref{TH:group} is hiding some finite-time dependency in $MK$ (possibly of the form of an extra-term in $\sum_{m,k}1/\Delta_{m,k}$, yet independent of $\log(T)$) that non-asymptotic algorithms\footnote{We call an algorithm non-asymptotic if its worst-case regret is of the order of $\sqrt{MKT}$, maybe up to some additional polynomial dependency in $M$ and $K$. In particular, classical algorithms for structured bandits \cite{combes2017minimal,Lattimore17} are only asymptotical.}
would not be able to remove.

\paragraph{Strong dominance} In the case of strong dominance, a similar closed-form expression can  be stated.
\begin{theorem}\label{Th:strong}
With strong dominance, a consistent algorithm verifies
$\displaystyle c_\mu = \sum_{k = 2}^K \frac{2}{\Delta_{1, k}} + \sum_{m =2}^M \frac{2}{\Delta_{m, K}} \,.$

\label{th:lower_strong}
\end{theorem}
This lower bound indicates that the dominance assumption can be leveraged to replace the asymptotic linear dependency in the total number of arms category into a linear dependency in the number of arms of the optimal category plus the number of categories. With $M$ categories of $K$ arms each, the dependency in $MK$ is replaced into $M+K$. However, as before and for the same reasons, the finite-time minimax lower bound will still be of the order $\sqrt{MKT}$.
The lower bound of Theorem \ref{Th:strong} seems to  indicate that an optimal algorithm should be pulling only the arms of the optimal category and the \textbf{worst} arm (not the best!) of the other categories, at least asymptotically  and logarithmically. Yet again, there is no guarantee that non-asymptotic algorithms can achieve this highly-demanding (and rather counter-intuitive) lower bound.

\paragraph{First-order dominance} There are no simple closed form expression of $c_\mu$ with the first-order dominance assumption, see nonetheless Appendix \ref{Sec:FOlowerProof} for some variational expression. However, for the sake of illustration, we provide a closed-form solution for a specific case.
\begin{theorem}
With first-order dominance and $M=K=2$  and assuming that arms are intertwined,
i.e.\ $\mu^1_1 > \mu^2_1 > \mu^1_2 > \mu^2_2$, a consistent algorithm satisfies
\begin{equation*}
c_\mu = \frac{2}{\Delta_{1, 2}} + \frac{2}{\Delta_{2, 2}} + \frac{2}{\Delta_{2,1}} \bigg( 1 - \frac{\left( \Delta_{2,2}-\Delta_{1,2} \right)^2}{\left( \Delta_{1,2} \right)^2 + \left( \Delta_{2,2} \right)^2} \bigg)  \,.
\end{equation*}

\label{th:lower_fosd}
\end{theorem}
It is quite interesting to compare this lower bound to the corresponding ones with group-sparsity where $c_\mu = \frac{2}{\Delta_{1,2}}$, with  strong dominance where  $c_\mu = \frac{2}{\Delta_{1,2}}+\frac{2}{\Delta_{2,2}}$  and without structure at all where $c_\mu = \frac{2}{\Delta_{1,2}}+\frac{2}{\Delta_{2,2}} + \frac{2}{\Delta_{2,1}}$. Clearly, lower bounds are, as expected, decreasing with additional structure. More interestingly, the first-order lower bound somehow interpolates between this two by multiplying the term   $\frac{2}{\Delta_{2,1}}$ by a factor $\rho \in (0,1)$; $\rho =0$ corresponding to the stronger assumption of strong dominance and $\rho= 1$ to the absence of dominance assumption.

\section{Algorithms and upper bounds}


\subsection{Optimism principle}

Our first algorithm is based on the principle of optimism in the face of uncertainty and is summarized in Algorithm~\ref{algo:GenericCatSE}. It behaves in three different ways depending on the number of categories that are called ``active''. The definition of an active category will depend on the assumption of dominance. Formally, let $\delta \in (0,1)$ be a confidence level (fixing the confidence level actually requires that the horizon $T$ is known, but there exist well understood anytime version of all these results \cite{degenne2016anytime}). At time step $t$, it computes the set of active categories, denoted $\mathcal{A}(t, \delta)$. The three states of Algorithm~\ref{algo:GenericCatSE} are then as follows:
\begin{wrapfigure}[11]{r}{0.45\linewidth}
\vspace{-11pt}
\centering
\begin{algorithm}[H]
\SetAlgoLined
 Pull each arm once \\
 \While{$t \leq T$}{
  Compute set of active categories $\mathcal{A}(t, \delta)$ \\
  \uIf{$|\mathcal{A}(t, \delta)| = 0$}{
    Pull all arms \\
  }
  \uElseIf{$|\mathcal{A}(t, \delta)| = 1$}{
    Perform \textsc{UCB}($\delta$) in the active category
  }
  \Else{
    Pull all arms in active categories \\
  }
 }
 \caption{\textsc{CatSE}($\delta$)} 
 \label{algo:GenericCatSE}
\end{algorithm}
\end{wrapfigure}
\begin{enumerate}
    \item $|\mathcal{A}(t, \delta)| = 0$: no category is active; the algorithm pulls all arms.
    \item $|\mathcal{A}(t, \delta)| = 1$: only one category is active; the algorithm performs \textsc{UCB}($\delta$) in it.
    \item $|\mathcal{A}(t, \delta)| > 1$: several categories are active; the algorithm pulls all arms inside those.
\end{enumerate}

We now detail what we called an active category for each notion of dominance defined previously along with theorems upper bounding the regret of the \textsc{CatSE} algorithm.

\paragraph{Group-sparse dominance} Under this assumption, we say a category is active if it has an active arm. Following the idea of sparse bandits~\cite{kwon2017sparse} or bounded regret \cite{bubeck2013bounded}, we say that the arm $k$ of category $m$ is active if
\begin{equation*}
    \textstyle \widehat{\mu}_k^m(t) := \frac{\sum_{s < t; (C_t, A_t)=(m,k)} X_{A_t}^{C_t} }{N_k^m(t)} \geq 2 \sqrt{\frac{\log N_k^m(t)}{N_k^m(t)}} \,.
\end{equation*}
This condition ensures that the expected number of times an arm with positive mean is non active is finite in expectation. Similarly,  the expected number of times an arm with non positive mean is active is also finite. Those conditions will ensure that the expected number of times a suboptimal category is pulled is also finite. Then, the set of active categories, denoted $\mathcal{A}(t)$ is simply
\begin{equation*}
    \textstyle \mathcal{A}(t) \coloneqq \bigg\{ m \in [M] ; \exists \, k \in [K], \widehat{\mu}_k^m(t)  \geq  2 \sqrt{\frac{\log N_k^m(t)}{N_k^m(t)}} \bigg\} \,.
\end{equation*}

\begin{theorem}
In the group-sparse dominance setting, the expected regret of \textsc{CatSE} verifies with probability at least $1-2\delta KT$,
\begin{equation*}
    \mathbb{E}[R_T] \leq \sum_{k=2 }^K \frac{8 \log\frac{1}{\delta}}{\Delta_{1,k}} + \sum_{m, k} \Delta_{m,k} + \frac{40}{(\mu_1^1)^2} \log \frac{16}{(\mu_1^1)^2} \sum_{m, k} \Delta_{m,k} + (M-1) K \frac{\pi^2}{6}  \sum_{m, k} \Delta_{m,k} \,.
\end{equation*}
\end{theorem}

The first term is the bound of the \textsc{UCB} algorithm while the third term is the regret incurred when the optimal category is non active and the last term comes from a suboptimal category being active. As a result, \textsc{CatSE} is asymptotically optimal, up to a multiplicative factor.
A trick to improve empirically the performance of the algorithm is to replace the round-robin sampling phase (when $|\mathcal{A}(t)| = 0$) by choosing an arm with a higher probability the closer it is to be active. This idea was analyzed in~\cite{bubeck2013bounded} with  additional assumptions. Yet this can only improve  the second term of the regret, which is already constant w.r.t.\ $T$ (so we chose to not focus on it). For example, a possibility is to pull arm $(m,k)$ at time $t$ with probability
$p^m_k(t) \propto \left( \sqrt{\frac{4 \log N^m_k(t)}{N^m_k(t)}} - \widehat{\mu}^m_k(t) \right)^{-2}$.
Another possible improvement is to eliminate categories in which there exist an arm whose upper bound is less than 0. Again, this only improves a term constant w.r.t.\ $T$.

\paragraph{Strong dominance} In this setting, \textsc{CatSE} will use the information gathered by all arms. The overall idea is to construct confidence region for the mean vector and to eliminate a category as soon as it is clearly dominated by another one. The statistical test to perform in order to determine which categories to eliminate is based on the following alternative characterization of dominance.

Let $\Delta(K) \coloneqq \{ \mathbf{x} \in \mathbb{R}^K_+; \| \mathbf{x} \|_1 = 1 \}$ be the $K$-simplex and $\mu^m \coloneqq (\mu_k^m)_k$ be the vector of means.

\begin{proposition}\label{PR:StrongDominance}
$\mathcal{A}$ strongly dominates $\mathcal{B}$ if and only if
$\forall \, \mathbf{x} \in \Delta(K), \forall \, \mathbf{y} \in \Delta(K), \langle \mathbf{x}, \mu^\mathcal{A} \rangle \geq \langle  \mathbf{y}, \mu^\mathcal{B} \rangle$.
\end{proposition}
At the end of the $p$-th round of the phase of successive elimination of categories, each arm  has been pulled  $p$ times. A natural estimator  of $\mu^m \in \mathbb{R}^K$ is  the coordinate wise empirical average of rewards, i.e., $\mu^m_k(p) = \frac{1}{p}\sum_{r=1}^p X^m_k(r)$, where (with a slight abuse of notation), $X^m_k(r)$ is the reward gathered by the $r$-th pull of arm $k$ of category $m$.
We  now describe the statistical run at the end of round $p \in \mathbb{N}$; category $n \in [M]$ is eliminated by category $m \in [M]$ if it holds that 
\begin{equation}
    L_m^+(p, \delta) \coloneqq \max_{\mathbf{x} \in \Delta(K)} \langle \mathbf{x} , \widehat{\mu}^m(p) \rangle - \left\lVert \mathbf{x} \right\rVert_2  \beta(p,\delta) > \min_{\mathbf{y}\in \Delta(K)} \langle \mathbf{y}, \widehat{\mu}^n(p) \rangle + \left\lVert \mathbf{y} \right\rVert_2  \beta(p,\delta)  =:  L_n^-(p, \delta) \,,
    \label{eq:CatSE0}
\end{equation}
where $\beta(p,\delta) = \sqrt{ \frac{2}{p} \left( K \log 2 + \log \frac{1}{\delta} \right)}$.
The set of active categories is then define as follows
\begin{equation*}
\mathcal{A}(t, \delta) = \left\{ m \in [M] ; \forall \, n \ne m, L_n^+(t, \delta) \leq L_m^-(t, \delta) \right\} \,.
\end{equation*}

\begin{theorem}
In the strong dominance case, the regret of \textsc{CatSE} satisfies w.p.\ at least $1 - \delta MT$,
\begin{equation*}
    R_T \leq \sum_{k=2}^K \frac{8 \log\frac{1}{\delta} }{\Delta_{1,k}}  + \sum_{m,k} \Delta_{m,k} + 8\big(K \log 2  + \log\frac{1}{\delta} \big) \sum_{m=2}^M  \min_{\mathbf{x}, \mathbf{y} \in \Delta(K)}  \Big( \frac{\left\lVert \mathbf{x} \right\rVert_2 + \left\lVert \mathbf{y} \right\rVert_2}{\langle \mathbf{x} , \mu^1 \rangle - \langle \mathbf{y} , \mu^m \rangle} \Big)^2 \sum_{k=1}^K  \Delta_{m,k}
\end{equation*}
\end{theorem}

\paragraph{First-order dominance} \textsc{CatSE} will proceed with first-order dominance as with strong dominance, the major difference is the statistical test. Let us first characterize the notion of first-order dominance. 

\begin{proposition}
$\mathcal{A}$ first-order dominates $\mathcal{B}$ if and only if
$    \forall\, \mathbf{x} \in \Delta(K), \langle \mathbf{x}, \mu^\mathcal{A} \rangle  \geq  \langle \mathbf{x}, \mu^\mathcal{B} \rangle \,.$
\end{proposition}

The statistical test is then: category $n \in [M]$ is eliminated by category $m \in [M]$ at round $p$ if 
\begin{equation}
    D_{m,n}(p, \delta) \coloneqq \max_{\mathbf{x} \in \Delta(K)}  \frac{\langle \mathbf{x}, \widehat{\mu}^{m}_{\sigma}(p)  -  \widehat{\mu}^{n}_\tau(p) \rangle}{\lVert \mathbf{x} \rVert_2} > 2  \gamma(p, \delta) \,,
    \label{eq:CatSE1}
\end{equation}
where $\widehat{\mu}^{m}_{\sigma}(p)$ and $\widehat{\mu}^{n}_\tau(p)$ represent respectively the reordering of $\widehat{\mu}^m(p)$ and $\widehat{\mu}^n(p)$ in decreasing order and $\gamma(p, \delta) = \frac{1}{\sqrt{2p}} \big( \sqrt{K\log\frac{1}{\delta}} + \sqrt{1 + (K+1)\log K} \big)$. We emphasis the permutation is specific to both a category and a round. This statistical test yields the following set of active categories
\begin{equation*}
    \mathcal{A}(t, \delta) = \left\{ m \in [M] ; \forall\, n \ne m, D_{m,n}(t, \delta) \leq 2 \gamma(t, \delta) \right\} \,.
\end{equation*}
\begin{theorem}
Under the additional assumption that $X^m_k \in [0,1]$ for all category $m$ and arm $k$, in the first-order dominance, the regret of \textsc{CatSE} verifies with probability at least $1 - \delta MT$,
\begin{equation*}
    R_T \leq \sum_{k=2}^K \frac{8 \log \frac{1}{\delta}}{\Delta_{1,k}} + \sum_{m,k} \Delta_{m,k} + 16 \left( K\log\frac{1}{\delta} + K\log K + \log K + 1 \right) \sum_{m=2}^M \frac{\sum_{k=1}^K \Delta_{m,k}}{\| \mu^1 - \mu^m \|_2^2} \,.
\end{equation*}
\end{theorem}

\subsection{Bayesian principle}

\begin{wrapfigure}[7]{r}{0.41\linewidth}
\vspace{-11pt}
\centering
\begin{algorithm}[H]
  \SetAlgoLined
  \While{$t \leq T$}{
  Sample $\theta(t) \sim \Pi_{t-1}\left( \cdot | \mathcal{H}_d \right)$\\
  Pull $(C_t, A_t) \in \argmax_{(m,k)} \theta_k^m(t)$
  }
    \caption{\textsc{Murphy Sampling}} 
    \label{algo:ms}
\end{algorithm}
\end{wrapfigure}
The \textsc{Murphy Sampling} (\textsc{MS}) algorithm~\cite{kaufmann2018sequential} was originally developed in a pure exploration setting. Conceptually, it  is derived from \textsc{Thompson Sampling} (\textsc{TS})~\cite{thompson1933likelihood}, the  difference is that the sampling respects some inherent structure of the problem.
To define \textsc{MS}, we denote by $\mathcal{F}(t) = \sigma\left( A_1, X_1, \dots, A_t, X_t \right)$  the information available after $t$ steps and $\mathcal{H}_d$ the assumption of dominance considered. Let $\Pi_t = \mathbb{P}\left( \cdot | \mathcal{F}_t \right)$ be the posterior distribution of the means parameters after $t$ rounds. The algorithm samples, at each time step, from the posterior distribution $\Pi_{t-1}\left( \cdot | \mathcal{H}_d \right)$ and then pulls the best arm, which, by definition, is in the best category sampled at this time step. In comparison, \textsc{TS} would sample from $\Pi_{t-1}$ without taking into account any structure. 
To implement this algorithm, we use that independent conjugate priors will produce independent posteriors, making the posterior sampling tractable. The required assumption, i.e.\ the structure of our problem, is then attained using rejection sampling. We do not provide theoretical guarantees on its regret but we will illustrate empirically on simulated data that it is highly competitive compared to the other algorithms.

\section{Experiments}

In this section, we present numerical experiments illustrating the performance of the algorithms we have introduced. We also compare them with two families of algorithms. The first one is algorithms for the multi-armed bandit framework, namely \textsc{UCB}~\cite{auer2002finite} and \textsc{TS}~\cite{thompson1933likelihood}; they are agnostic to the structure of the arms. The second family of algorithms is adapted to tree search, namely \textsc{UCT}~\cite{kocsis2006bandit}; they partially take into account the inherent structure. Specifically, they will just use the fact that arms are grouped into categories but not that one category dominates the others. 
We consider two scenarios for the different dominance hypothesis. In all experiments, rewards are drawn from Gaussian distribution with unit variance and we report the average regret as a function of time, in log-scale. To implement \textsc{TS} and \textsc{MS}, we pulled each arm once and then sampled using a Gaussian prior. The simulations were ran until time horizon 10,000 and results were averaged over 100 independent runs.

\begin{figure}[t]
\centering
\begin{subfigure}{0.49\textwidth}
\centering
\includegraphics[width=1\linewidth]{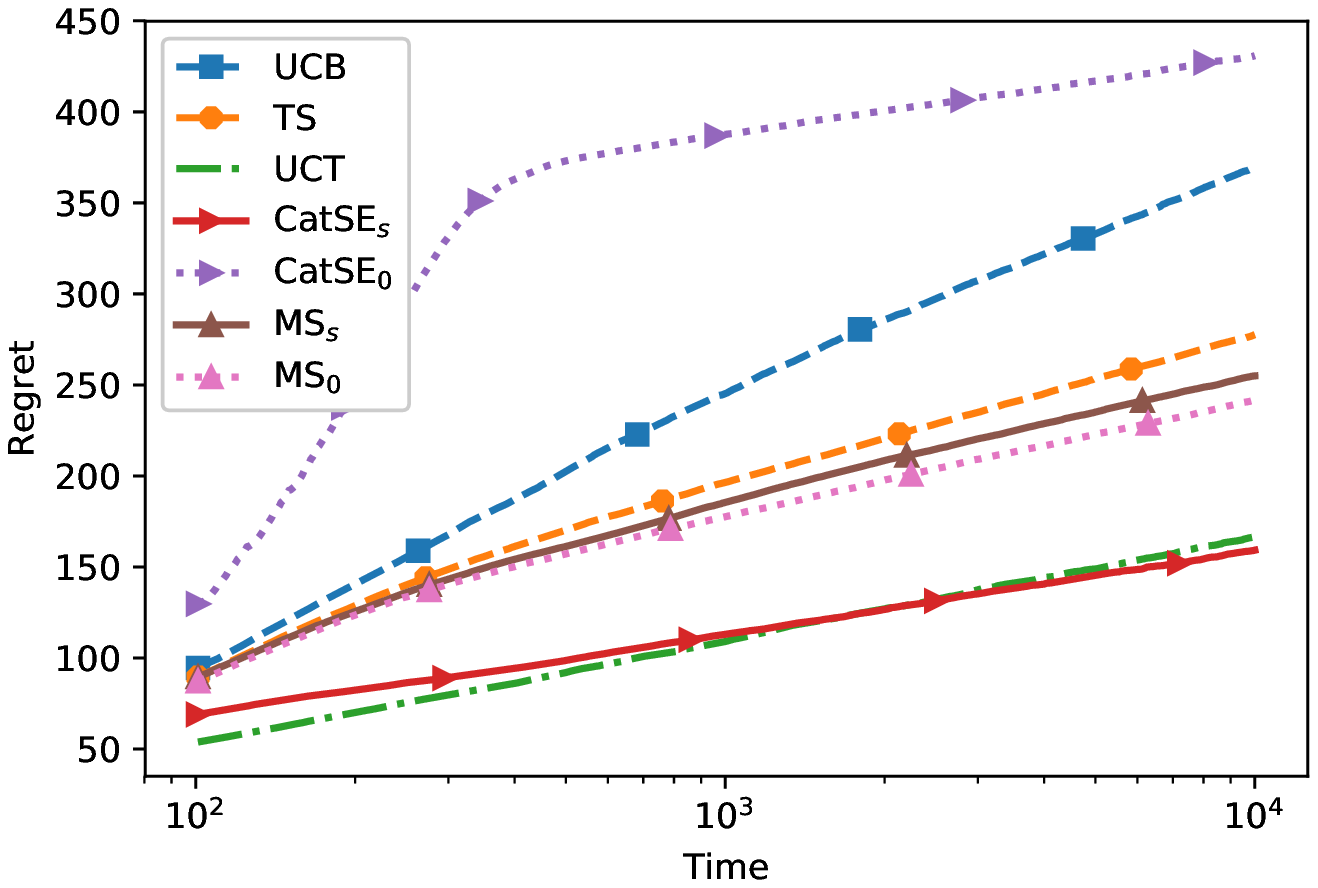}
\caption{In the sparse and strong dominance scenario}
\label{fig:sparse-strong}
\end{subfigure}
\begin{subfigure}{0.49\textwidth}
\centering
\includegraphics[width=1\linewidth]{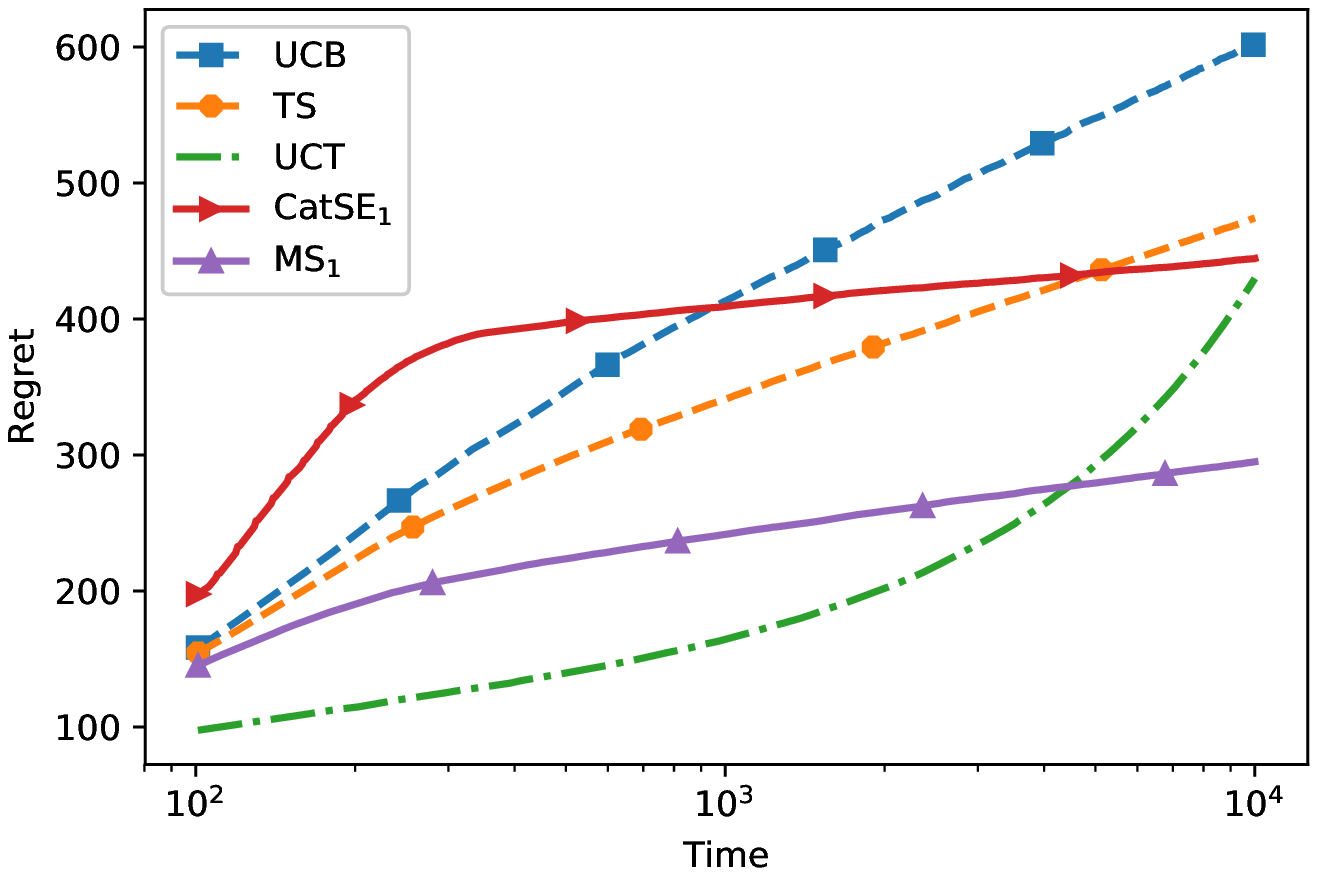}
\caption{In the first-order dominance scenario}
\label{fig:fod}
\end{subfigure}
\caption{Regret of various algorithms as a function of time}
\label{fig:exp}
\end{figure}

\paragraph{Group-sparse \& strong dominance} We start by grouping the experiments in the group-sparse and strong dominance setting, as we recall that the only difference between the two concepts is the knowledge of a threshold between the best category and the others. In this first scenario, we analyze a problem with five categories and five arms per category. Precisely, in the first category the optimal arm has expected reward $1$, and the four suboptimal arms consist of one group of three (stochastically) identical arms each with expected reward $0.5$ and one arm with expected reward $0$. The four suboptimal category are identical and are composed of two arms with expected rewards $0$ and $-1$, respectively and a group of three arms with expected reward $-0.5$. We used the subscript $s$ and $0$ to denote the assumption of dominance the algorithm exploited. $\textsc{CatSE}_s$ and $\textsc{CatSE}_0$ were run with $\delta = \frac{1}{t}$ and $\delta = \frac{1}{Mt}$, respectively. Results are presented on Figure~\ref{fig:sparse-strong}.
In the case of group-sparse dominance, $\textsc{CatSE}_s$ 
outperforms both \textsc{UCB} and \textsc{UCT}; $\textsc{MS}_s$ asymptotically performs as well yet with a slightly higher regret. Interestingly, \textsc{UCT} performs well in the beginning; thanks to the lack of an exploration phase compared to $\textsc{CatSE}_s$.
In the case of strong dominance, $\textsc{MS}_0$ and $\textsc{CatSE}_0$ asymptotically perform alike and slightly better than \textsc{UCT}. However, the regret of $\textsc{CatSE}_0$ is much higher due to its round-robin sampling phase; this can be seen in the beginning as $\textsc{CatSE}_0$ is still in the search of the optimal category.
If we compare the two versions of each algorithm between them, we can notice two points. Firstly, for \textsc{CatSE}, the result of the potential sampling improvement is significant. Secondly, for \textsc{MS}, the regret in the group-sparse case is slightly worse than in the strong dominance case even though it is stronger. This is simply due to our implementation and the difficulty of the posterior sampling, in particular the rejection sampling phase.

\paragraph{First-order dominance} 
Finally, we consider the first-order dominance setting. In this scenario, we look upon a problem with five categories and ten arms per category. Precisely, in the optimal category, the best arm has expected reward 5 while the nine suboptimal arms consist of three group of five, three and one arms, with expected rewards $4$, $3$ and $2$, respectively. The four suboptimal categories are composed of two arms with expected rewards $4.5$ and $0$, respectively, and eight arms with expected reward $3$. \textsc{CatSE} was run with $\delta = \frac{1}{Mt}$ and the results are presented on Figure~\ref{fig:fod}.
Once again, \textsc{MS} and \textsc{CatSE} outperform baseline algorithms and both appear to have the same slope asymptotically with a significant difference between their regret, again due to the exploration phase of \textsc{CatSE}. It is interesting to observe that \textsc{UCT} performed poorly; as noticed in~\cite{coquelin2007bandit}, the convergence can be sluggish. Indeed, the main issue occurs when the best arm is underestimated. In that case, it is pulled a logarithmic number of times the optimal category is pulled, which is a logarithmic number of times, since the second best arm overall is in suboptimal categories. Hence, it would take an exponential of exponentials number of time for the optimal arm to become the best again.

\section{Conclusion}


Two problems remain open: the first one is a better exploration phase in \textsc{CatSE} since it heavily impacts the regret and as noted in~\cite{garivier2016explore}, ETC algorithms are necessarily suboptimal; 
and the second is an upper bound on the regret of the \textsc{MS} algorithm since it is highly competitive in practice. We believe that it is asymptotically optimal and that it can be applied to other setting of structured bandits.

\subsubsection*{Acknowledgments}

This work was supported in part by a public grant as part of the Investissement d'avenir project, reference ANR-11-LABX-0056-LMH, LabEx LMH, in a joint call with Gaspard Monge Program for optimization, operations research and their interactions with data sciences.

\medskip


\bibliographystyle{plainnat}
\bibliography{references}

\newpage

\appendix

\section{Proofs of lower bounds}

\subsection{Group-sparse dominance}

The set $\Lambda(\mu)$ in the optimization problem can be decomposed into $\Lambda(\mu) = \Lambda_k(\mu) \sqcup \dots \sqcup \Lambda_K(\mu)$ where $\Lambda_k(\mu)$ is the set of alternative parameters in which arm $k$ of category 1 is optimal. Indeed, as we know that $\lambda_1^1=\mu_1^1 > 0$, the best category is known and the regret incurred by suboptimal categories is non-existent. Thus, asymptotically, we fall back into deriving a lower bound on the regret in one category, i.e.\ in the classic multi-armed bandit setting. 

\subsection{Strong dominance}

Without loss of generality, we assume that we have $M=2$ categories and category 2 has a unique worst arm. The condition in the optimization problem can be written as
\begin{equation*}
    \sum_{k =2}^K N_k^1 (\mu_k^1 - \lambda_k^2)^2 + \sum_{k=1}^K N_k^2 (\mu_k^2 - \lambda_k^2)^2 \geq 2, \forall\, \lambda \in \Lambda(\mu) \,,
\end{equation*}
where $\Lambda(\mu) = \Lambda_2(\mu) \sqcup \dots \sqcup \Lambda_{K}(\mu) \sqcup \Lambda^2(\mu)$ where $\Lambda_k(\mu)$ is the event in which the best arm is mistaken by arm $k$ in the category 1, i.e.,
\begin{equation*}
    \Lambda_k(\mu) = \{\mu_1^1\} \times ]-\infty, \mu_1^1[ \times \dots \times ]\mu_1^1, +\infty[ \times \dots \times ]-\infty, \mu_1^1[ \times ]-\infty, \mu_1^1[ \times \dots \times ]-\infty, \mu_1^1[
\end{equation*}
and $\Lambda^2(\mu)$ is the event in which we mistake category 2 as the optimal category, i.e.,
\begin{equation*}
    \Lambda^2(\mu) = \{\mu_1^1\} \times ]-\infty, \mu_1^1[ \times \dots \times ]-\infty, \mu_1^1[ \times ]\mu_1^1, +\infty[ \times \dots \times ]\mu_1^1, +\infty[ \,.
\end{equation*}
On $\Lambda_k(\mu)$, the condition is equivalent to
\begin{equation*}
    N_k^1 \left(\mu_1^1 - \mu_k^2 \right)^2 \geq 2 \,,
\end{equation*}
and on $\Lambda^2(\mu)$,
\begin{equation*}
    \sum_{k = 1}^K N_k^2 \left(\mu_1^1 - \mu_k^2 \right)^2 \geq 2 \,.
\end{equation*}
The minimization problem can thus be separated in two parts: the first part corresponds to finding the best arm in the optimal category and the second part to finding the optimal category.

For the first part, the solution is the same as in the multi-armed bandit setting and is given by $N^1_k = \frac{2}{\left(\Delta_{1,k}\right)^2}$. 

For the second part, let us prove that the solution is given by $N^2_K = \frac{2}{\left(\Delta_{2,K}\right)^2}$ and $N^2_k = 0$ for $k \ne K$. We have the following problem
\begin{equation*}
    \min_{N^2 \geq 0} \sum_{k=1}^K N_k^2 \Delta_{2,k} =: f(N^2) \qquad \text{subject to } \sum_{k=1}^K N_k^2 \left(\Delta_{2,k}\right)^2 \geq 2 \,.
\end{equation*}
On one side, we have
\begin{equation*}
    \min_{N \geq 0} f(N) \leq \min_{n \geq 0} f(0, \dots, 0, n) = f\left(0, \dots, 0,  \frac{2}{\left(\Delta_{2,K}\right)^2} \right) = \frac{2}{\Delta_{2,K}} \,,
\end{equation*}
and on the other side, since $\Delta_{2,k} < \Delta_{2,K}$, we have
\begin{equation*}
    \sum_{k=1}^K N_k^2 \Delta_{2,k} > \frac{1}{\Delta_{2,K}} \sum_{k=1}^K N_k^2 \left( \Delta_{2,k} \right)^2 \geq \frac{2}{\Delta_{2,K}} \,.
\end{equation*}
Hence the solution of the optimization problem in the suboptimal category and the lower bound on the regret follows.

\subsection{First-order dominance}\label{Sec:FOlowerProof}

By simplifying the optimization problem, one obtains the two following conditions

\begin{equation*}
    \forall\, k \ne 1, N^1_k \left( \Delta_{1, k} \right)^2 \geq 2 \,,
\end{equation*}
and $\forall\, k \in [K]$,
\begin{equation*}
    \begin{split}
        &\sum_{i=1}^{k-1} \left[ \left( N^1_{i+1} \left( \mu^1_{i+1} - \widetilde{\mu}_i \right)^2 + N^2_i \left( \mu^2_i - \widetilde{\mu}_i \right)^2 \right) \mathbf{1}\left\{ \mu^2_i < \mu^1_{i+1} \right\} \right] + N^2_k \left( \Delta_{2, k} \right)^2 \\
        &+ \sum_{j=k+1}^K \left( N^1_j \left( \mu^1_j - \overline{\mu}_j \right)^2 + N^2_j \left( \mu^2_j - \overline{\mu}_j \right)^2 \right) \geq 2 \,,
    \end{split}
\end{equation*}
where $\widetilde{\mu}_i = \frac{N^1_{i+1} \mu^1_{i+1} + N^2_i \mu^2_i}{N^1_{i+1} + N^2_i}$ and $\overline{\mu}_j = \frac{N^1_j \mu^1_j + N^2_j \mu^2_j}{N^1_j + N^2_j}$.

Assuming the arms are intertwined, the first term in the above equation disappear since the condition in the indicator function is not verified. In the case of $M=2$ categories and two arms per category $K=2$, the following conditions are derived
\begin{equation*}
    N^1_2 \geq \frac{2}{\left( \Delta_{1,2} \right)^2} \text{,} \qquad N^2_2 \geq \frac{2}{\left( \Delta_{2,2} \right)^2} \,,
\end{equation*}
and 
\begin{equation*}
    N^2_1 \left( \Delta_{2,1} \right)^2 + N^1_2 \left( \mu^1_2 - \overline{\mu} \right)^2 + N^2_2 \left( \mu^2_2 - \overline{\mu} \right)^2 \geq 2 \,,
\end{equation*}
where $\overline{\mu} = \frac{N^1_2 \mu^1_2 + N^2_2 \mu^2_2}{N^1_2 + N^2_2}$.

Since this is a minimization problem, it is clear that the regret is minimize on the lower bounds of $N^1_2$ and $N^2_2$. Putting this two quantities in the last inequality, we obtain
\begin{equation*}
    N^2_1 \geq \frac{2}{\left( \Delta_{2,1} \right)^2} \left[ 1 - \left( \left( \frac{\mu^1_2 - \overline{\mu}}{\Delta_{1,2}} \right)^2 + \left( \frac{\mu^2_2 - \overline{\mu}}{\Delta_{2,2}} \right)^2 \right) \right] \,.
\end{equation*}
Developing $\overline{\mu}$, we have
\begin{equation*}
    \overline{\mu} = \frac{\frac{2 \mu^1_2}{\left( \Delta_{1,2} \right)^2} + \frac{2 \mu^2_2}{\left( \Delta_{2,2} \right)^2}}{\frac{2}{\left( \Delta_{1,2} \right)^2} + \frac{2}{\left( \Delta_{2,2} \right)^2}} = \frac{\mu^1_2 \left( \Delta_{2,2} \right)^2 + \mu^2_2 \left( \Delta_{1,2} \right)^2}{\left( \Delta_{1,2} \right)^2 + \left( \Delta_{2,2} \right)^2} \,.
\end{equation*}
Now developing $\frac{\mu^1_2 - \overline{\mu}}{\Delta_{1,2}}$, we get:
\begin{equation*}
    \frac{\mu^1_2 - \overline{\mu}}{\Delta_{1,2}} = \frac{\Delta_{1,2} \left( \mu^1_2 - \mu^2_2 \right)}{\left( \Delta_{1,2} \right)^2 + \left( \Delta_{2,2} \right)^2} = \frac{\Delta_{1,2} \Delta^{1,2}_{2,2}}{\left( \Delta_{1,2} \right)^2 + \left( \Delta_{2,2} \right)^2} \,.
\end{equation*}
Similarly,
\begin{equation*}
    \frac{\mu^2_2 - \overline{\mu}}{\Delta_{2,2}} = - \frac{\Delta_{2,2} \Delta^{1,2}_{2,2}}{\left( \Delta_{1,2} \right)^2 + \left( \Delta_{2,2} \right)^2} \,.
\end{equation*}
Plugging this into the inequality on $N^2_1$, we obtain
\begin{equation*}
    N^2_1 \geq \frac{2}{\left( \Delta_{2,1} \right)^2} \left[ 1 - \frac{\left(\Delta^{1,2}_{2,2}\right)^2}{\left( \Delta_{1,2} \right)^2 + \left( \Delta_{2,2} \right)^2} \right] \,.
\end{equation*}
The result follows by the decomposition of the expected regret.

\section{Characterizations of dominance}

\subsection{Strong dominance}

Let $(e_i)_i$ denotes the unit vectors. Taking $\mathbf{x} = e_k$ and $\mathbf{y} = e_l$ hands $\mu^1_k \geq \mu^2_l$.

In the other direction, let $(\alpha$, $\beta) \in \Delta(K) \times \Delta(K)$. We have
\begin{equation*}
    \langle \alpha,  \mu \rangle = \sum_{k = 1}^K \alpha_k \mu_k = \sum_{k = 1}^{K-1} \alpha_k \mu_k + \left( 1 - \sum_{k = 1}^{K-1} \alpha_k \right) \mu_K = \mu_K + \sum_{k = 1}^{K-1} \alpha_k ( \mu_k - \mu_K ) \,.
\end{equation*}
Now, using the previous equality, we obtain
\begin{equation*}
    \langle \alpha, \mu^1 \rangle - \langle \beta, \mu^2 \rangle = \sum_{k = 1}^K \alpha_k \mu_k^1 - \sum_{k = 1}^K \beta_k \mu_k^2 = (\mu_K^1 - \mu_1^2) + \sum_{k = 1}^{K-1} \alpha_k ( \mu_k^1 - \mu_K^1 ) + \sum_{k = 2}^{K} \beta_k ( \mu_1^2 - \mu_k^2 ) \geq 0 \,.
\end{equation*}

\subsection{First-order dominance}

Taking $\mathbf{x} = e_k$ hands $\mu^1_k \geq \mu^2_k$. In the other direction, let $\mathbf{x} \in \Delta(K)$. We have
\begin{equation*}
    \langle \mathbf{x} , \mu^1 - \mu^2 \rangle = \sum_{k=1}^K \mathbf{x}_k (\mu^1_k - \mu^2_k) \geq 0 \,.
\end{equation*}


\section{Regret upper bounds of \textsc{CatSE}}

\subsection{Group-sparse dominance}

Consider the following clean event
\begin{equation*}
    \mathcal{E}_s = \left\{ \forall\, t \in [T], \forall\, k \in [K], |\widehat{\mu}^1_k(t) - \mu^1_k| \leq \sqrt{\frac{2 \log \frac{1}{\delta}}{N^1_k(t)}} \right\} \,.
\end{equation*}
Using union bounds over $t$ and $k$, one obtains thanks to the subGaussian assumption that $\mathbb{P}\left(\mathcal{E}_s\right) \geq 2 \delta KT$. In the following, we assume the clean event holds true. In the case in which only the optimal category is active, we get the regret of the \textsc{UCB} algorithm
\begin{equation*}
    R_T \leq  \sum_{k=2}^K \frac{8 \log\frac{1}{\delta}}{\Delta_{1,k}} \,.
\end{equation*}

On the other hand, the set of active categories is empty if the optimal category is non active. That means that $\forall\, k \leq s, \widehat{\mu}_k^1(N_k^1(t)) < 2 \sqrt{\frac{\log N_k^1(t) }{N_k^1(t)}}$ where $s$ is the number of arms with positive expected reward. Let $\mathcal{A}_s$ denote this event. The number of times it happen is bounded. Indeed, since
\begin{equation*}
    \mathcal{A}_s  \subseteq \left\{ \widehat{\mu}_1^1(N_1^1(t)) < 2 \sqrt{\frac{\log N_1^1(t) }{N_1^1(t)}} \right\} =: \mathcal{A}_1 \,,
\end{equation*}
and
\begin{equation*}
    n \geq 3 + \frac{32}{(\mu_1^1)^2} \log \frac{16}{(\mu_1^1)^2} \Rightarrow 2 \sqrt{\frac{\log n}{n}} - \mu_1^1 \leq - \frac{\mu_1^1}{2} \,,
\end{equation*}
we have
\begin{align*}
    \mathbb{E}\left[ \sum_{t=MK+1}^T \mathbf{1}\left\{ \mathcal{A}_s \right\} \right] &\leq \mathbb{E}\left[ \sum_{t=MK+1}^T \mathbf{1}\left\{ \mathcal{A}_1 \right\} \right] \\
    &\leq \left( 3 + \frac{32}{(\mu_1^1)^2} \log \frac{16}{(\mu_1^1)^2} \right) + \sum_{u=1}^T \mathbb{P}\left( \widehat{\mu}_1^1(u) - \mu_1^1 < - \frac{\mu_k^1}{2} \right) \\
    &\leq \left( 3 + \frac{32}{(\mu_1^1)^2} \log \frac{16}{(\mu_1^1)^2} \right) + \sum_{u=1}^T \exp\left\{ - \frac{u}{8} (\mu_1^1)^2 \right\} \\
    &\leq 3 + \frac{32}{(\mu_1^1)^2} \log \frac{16}{(\mu_1^1)^2}  + \frac{8}{(\mu_1^1)^2} \,.
\end{align*}

Finally, the set of active categories has more than one element if a sub-optimal category is active, i.e. $\exists\, m \ne 1, \exists\, k \in [K]; \widehat{\mu}_k^m(N_k^m(t)) \geq 2 \sqrt{\frac{\log N_k^m(t)}{N_k^m(t)}}$. Let $\mathcal{B}$ denote this event. The number of times it happen is also bounded. Indeed,
\begin{align*}
    \mathbb{E} \sum_{t=1}^T \mathbf{1}\left\{ \mathcal{B} \right\} &\leq \sum_{m, k} \sum_{u=1}^T \mathbb{P}\left( \widehat{\mu}_k^m(u) \geq 2 \sqrt{\frac{\log u}{u}} \right) \\
    &\leq \sum_{m, k} \sum_{u=1}^T \mathbb{P}\left( \widehat{\mu}_k^m(u) - \mu_k^m \geq 2 \sqrt{\frac{\log u}{u}} \right) \\
    &\leq \sum_{m, k} \sum_{u=1}^T \frac{1}{u^2} \leq (M-1) K \frac{\pi^2}{6} \,.
\end{align*}

Combining the three inequalities, we conclude.

\subsection{Strong dominance}

Let $\mathcal{E}_0$ denote the clean event
\begin{equation*}
    \mathcal{E}_0 = \left\{ \forall\, t \in [T] ;  \forall\, m \in [M], \forall \mathbf{x}\, \in \mathbb{R}^K, \langle \mathbf{x}, \widehat{\mu}^m(t) - \mu^m \rangle \leq \lVert \mathbf{x} \rVert_2 \beta(t, \delta) \right\} \,,
\end{equation*}
where $\beta(t, \delta)=\sqrt{\frac{2}{N^m(t)} \left(K \log 2 + \log\frac{1}{\delta} \right)}$. 

\begin{lemma}
With probability at least $1 - \delta$, the following holds uniformly overall all $ \mathbf{x} \in \mathbb{R}^K$,
\begin{equation*}
    \langle \mathbf{x}, \widehat{\mu}^m(p) - \mu^m \rangle \leq  \left\lVert \mathbf{x} \right\rVert_2  \sqrt{ \frac{2}{p} \left( K \log 2 + \log\frac{1}{\delta} \right)} \,.
\end{equation*}
\end{lemma}
\begin{proof}
Fix $x \in \mathbb{R}$ and $\delta \in (0,1)$ a confidence level. According to (Lattimore and Szepesv{\'a}ri, 2018), we have with probability at least $1 - \delta$,
\begin{equation*}
    \left\lVert \widehat{\mu}(t) - \mu \right\rVert_{V_t} \leq  \sqrt{2 \left( K \log 2 + \log \frac{1}{\delta} \right)} \,.
\end{equation*}

If an agent pulls each arm sequentially, we are in the fixed design setting. In this case, (assuming $t$ is a multiple of $K$), we have $V_t = N(t) \mathbf{I}_K$, i.e.\ it is a diagonal matrix and we conclude.
\end{proof}

Using union bounds over the time and the categories, and using the definition of the confidence set, we obtain $\mathbb{P}\left(\mathcal{E}_0^c \right) \leq \delta MT$. 

Suppose we are in the clean event and let $m \ne 1$ and $t$ be the last time when we did not invoke the stopping rule, i.e.\ that the category $m$ is still active. First remark that category $1$ is never eliminated by category $m$ on the clean event since $\min_{k} \mu^1_k \geq \max_k \mu^m_k$. By Equation~\eqref{eq:CatSE0}, this means that
\begin{equation*}
\forall\, \mathbf{x} \in \Delta(K), \forall\, \mathbf{y} \in \Delta(K), \langle \mathbf{x}, \widehat{\mu}^1(t)\rangle - \langle \mathbf{y}, \widehat{\mu}^m(t)\rangle \leq \left( \| \mathbf{x} \|_2 + \| \mathbf{y} \|_2 \right) \sqrt{\frac{2}{N(t)}\left( \log \frac{1}{\delta} + K \log 2 \right)} \,,
\end{equation*}
where $N(t)$ denotes the number of times each category have been pulled. As we are in the clean event, we have
\begin{equation*}
\forall\, \mathbf{x} \in \Delta(K), \forall\, \mathbf{y} \in \Delta(K), \langle \mathbf{x}, \mu^1\rangle - \langle \mathbf{y}, \mu^m\rangle \leq 2 \left( \| \mathbf{x} \|_2 + \| \mathbf{y} \|_2 \right) \sqrt{\frac{2}{N(t)}\left( \log \frac{1}{\delta} + K \log 2 \right)} \,.
\end{equation*}
Inverting this equation, we obtain the following upper bound on $N(t)$
\begin{equation*}
    \forall\, \mathbf{x} \in \Delta(K), \forall\, \mathbf{y} \in \Delta(K), N(t) \leq 8 \left( K\log 2 + \log \frac{1}{\delta} \right) \left( \frac{\left\lVert \mathbf{x} \right\rVert_2 + \left\lVert \mathbf{y} \right\rVert_2}{\langle \mathbf{x} , \mu^1 \rangle - \langle \mathbf{y} , \mu^m \rangle} \right)^2 \,.
\end{equation*}

The proof is conclude with the proof of the \textsc{UCB} algorithm~\cite{auer2002finite}.
 
\subsection{First-order dominance}

\begin{lemma} With probability at least $1- \delta$,
\begin{equation*}
    \| \widehat{\mu}^m_{\sigma^m_t}(t) - \mu^m \|_2 \leq \frac{1}{\sqrt{2t}} \left( \sqrt{K\log\frac{1}{\delta}} + \sqrt{1 + (K+1)\log K} \right) \,,
\end{equation*}
where $\widehat{\mu}^m_{\sigma^m_t}(t)$ denotes the vector $\widehat{\mu}^m(t)$ ordered in decreasing order.
\end{lemma}
\begin{proof}
The McDiarmid inequality gives the following
\begin{equation*}
    \mathbb{P}\Big\{\|\widehat{\mu}^m_{\sigma^m_t}(t)-\mu^m\| \geq \mathbb{E} \|\widehat{\mu}^m_{\sigma^m_t}(t)-\mu^m\| + \varepsilon  \Big\} \leq \exp(-2t\varepsilon^2/K)
\end{equation*}
Now we just has to bound $\mathbb{E}\|\widehat{\mu}^m_{\sigma^m_t}(t)-\mu^m\|_2$. If $Y_1,\ldots, Y_N$ are $\sigma^2$ sub-Gaussian, then
\begin{equation*}
    \mathbb{P} \left\{\max_{i=1,\ldots,N} Y_i \geq \varepsilon \right\} \leq N \exp\left(-\frac{\varepsilon^2}{2\sigma^2}\right).
\end{equation*}
This give, by a careful integration, that 
\begin{equation*}
    \mathbb{E}\left(\max_{i=1,\ldots, N} Y_i\right)^2 \leq 2\sigma^2 (\log(N)+1)\,.
\end{equation*}
In our case, we have $\sigma^2 = \frac{1}{4t}$. Using that the expectation of the $k$\textsuperscript{th} maximum of $N$ random variables is smaller than the expectation of the maximum of $N-(k-1)$ random variables~\cite{david2004order}, we obtain
\begin{equation*}
    \mathbb{E}\| \widehat{\mu}^m_{\sigma^m_t}(t) - \mu^m \|_2^2 \leq \frac{1}{2t} \sum_{k=1}^K \left( 1 + \log(K-(k-1)) \right) = \frac{1}{2t} (K + \log K!) \leq \frac{1+(K+1) \log K}{2t} \,,
\end{equation*}
where the last inequality comes from the Stirling formulae. The result follows.
\end{proof}

Let define the clean event
\begin{equation*}
    \mathcal{E}_1 = \left\{ \forall\, t \in [T], \forall\, m \in [m], \| \widehat{\mu}^m_{\sigma^m_t}(t) - \mu^m \|_2 \leq \frac{1}{\sqrt{2t}} \left( \sqrt{K\log\frac{1}{\delta}} + \sqrt{1 + (K+1)\log K} \right) \right\}
\end{equation*}
By the lemma and with union bounds over $t$ and $m$, we have $\mathbb{P}\left( \mathcal{E}_1^c \right) \leq \delta MT$. Let $m \ne 1$ and $t$ be the last time we pulled category $m$.

By Equation~\eqref{eq:CatSE1}, we have
\begin{equation*}
    \forall\, \mathbf{x} \in \Delta(K), \langle \mathbf{x}, \widehat{\mu}^1_{\sigma^1_t}(t) - \widehat{\mu}^m_{\sigma^m_t}(t) \rangle \leq 2 \| \mathbf{x} \|_2 \gamma(t, \delta) \,.
\end{equation*}
Moreover, notice that after $t$ samples
\begin{align*}
    \forall\, \mathbf{x} \in \Delta(K), \frac{1}{\|\mathbf{x}\|_2}\Big| \langle \mathbf{x}, \widehat{\mu}^1_{\sigma^1_t}(t) - \widehat{\mu}^m_{\sigma^m_t}(t) \rangle - \langle \mathbf{x}, {\mu}^1 - {\mu}^m \rangle \Big| &\leq  \|\widehat{\mu}^1_{\sigma^1_t}(t)-\mu^1\|_2 + \|\widehat{\mu}^m_{\sigma^m_t}(t)-\mu^m\|_2 \\
    &\leq 2 \gamma(t, \delta) \,,
\end{align*}
where the last inequality holds true with probability at least $1-\delta MT$. Combining the two inequalities, one obtains with probability at least $1-\delta MT$,
\begin{align*}
    N^m(t) &\leq \frac{8}{\| \mu^1 - \mu^m \|_2^2} \left( \sqrt{K\log \frac{1}{\delta}} + \sqrt{1 + (K+1)\log K} \right)^2 \\
    &\leq \frac{16}{\| \mu^1 - \mu^m \|_2^2} \left( K\log\frac{1}{\delta} + K\log K + \log K + 1 \right)
\end{align*}
where in the last inequality we used the Cauchy–Schwarz inequality. Hence the result.




\end{document}